# Financial Instruction Following Evaluation (FIFE)

**Glenn Matlin**[1 2 3] **Siddharth**[1] **Anirudh JM**[2] **Aditya Shukla**[3] **Yahya Hassan**[3 2]
**Sudheer Chava**[1 2 3]

[1]Financial Services Innovation Lab, Georgia Institute of Technology
[2]College of Business, Georgia Institute of Technology
[3]College of Computing, Georgia Institute of Technology

glenn@gatech.edu siddharth889@gmail.com amahesh64@gatech.edu
ashukla73@gatech.edu yhassan30@gatech.edu sudheer.chava@scheller.gatech.edu

## Abstract

Language Models (LMs) struggle with complex, interdependent instructions, particularly in high-stakes domains like finance where precision is critical. We introduce FIFE, a novel, high-difficulty benchmark designed to assess LM instruction-following capabilities for financial analysis tasks. FIFE comprises 88 human-authored prompts and employs a verification system with chainable, verifiable constraints for fine-grained reward signals. We evaluate 53 models (proprietary, open-weight, open-source) in a zero-shot setting. Our key findings reveal a clear performance hierarchy: the top open-weight model (76.1 strict / 79.5 loose) surpasses the leading proprietary system (65.9 strict / 70.5 loose), while the best open-source models lag significantly (45.5 strict / 48.9 loose). However, even top-performing models struggle with FIFE's complex requirements, failing to achieve perfect compliance. We release our dataset and code as an open-source resource to promote research in Reinforcement Learning for the financial domain.[1].

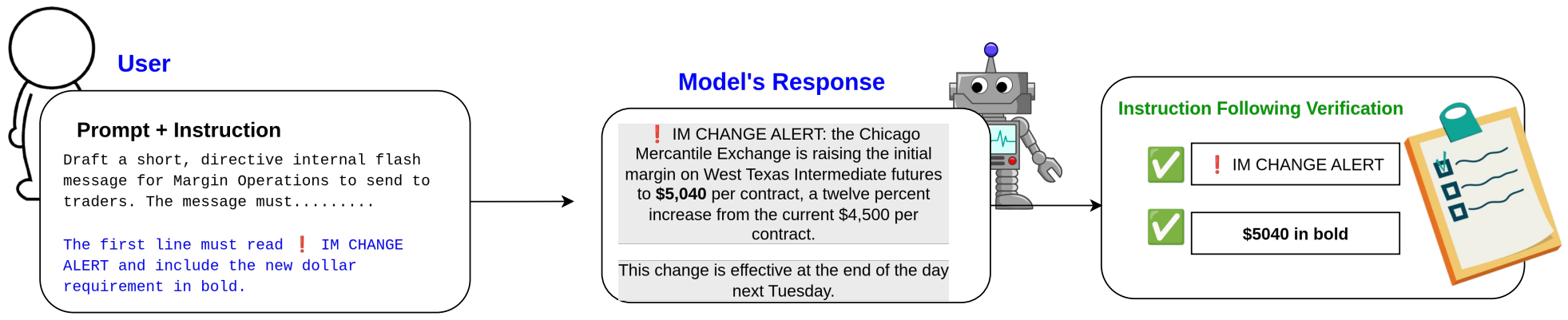

Figure 1: Instruction Following Pipeline.

## 1 Introduction

The value of Language Models (LMs) relies on their ability to follow instructions precisely [1], [2]. While instruction tuning has improved model obedience [3], [4], recent benchmarks reveal a critical gap: LMs still fail to adhere to long or complex instructions that require decomposing interdependent constraints [5], [6], [7], [8], [9].

This gap is particularly acute in high-stakes domains like finance. Here, workflows combine domain reasoning with strict, machine-checkable constraints; for example, *"Create a table with exact headers, sort by spread descending, and bold values above a threshold."* A factually correct answer that violates formatting instructions can cause operational failures. While existing Financial NLP

[1]The code and data are available for review at: https://github.com/gtfintechlab/FIFE

(FinNLP) benchmarks measure task-specific accuracy for tasks like sentiment analysis or QA [10], [11], [12], [13], they do not evaluate a model's ability to adhere to the complex, multi-part constraints common in financial queries or handle multi-turn conversational contexts [14].

To address this combination of compositional and structural constraints, we introduce Finance Instruction Following Evaluation (`FIFE`), a benchmark that stress-tests LMs using 88 human-authored tasks mirroring realistic financial workflows. Its verifiable instructions cover a wide range of compositional and stylistic constraints such as table structures, conditional formatting, and domain-specific data normalization. Using this framework, we evaluate 53 leading proprietary and open-source LMs in a zero-shot setting, scoring compliance automatically via regex-based verification. We release the benchmark, including all prompts, verifiers, and evaluation code, as an open-source tool to support future research into verifiable-reward training for high-stakes domains.

Our contributions are fourfold: (1) we introduce `FIFE`, the first instruction-following benchmark for finance, with 88 human-authored prompts requiring verifiable, multi-step reasoning; (2) we propose a robust evaluation protocol using strict and loose scoring with a registry of 88 chainable constraints; (3) we evaluate 53 models, revealing that open-weight systems can match or exceed proprietary model performance, though all struggle with complex compliance; and (4) we release our framework as an open-source toolkit to encourage research in verifiable-reward training.

| **Composition Type** | Description | Example | Illustration |
|---|---|---|---|
| Single | The output satisfies a single constraint or requirement | Write a 200-word end-of-day crypto market recap. Include Bitcoin's seven-day realized volatility decline from 45% to 32%. | 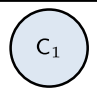 |
| And | The output satisfies multiple constraints simultaneously | Write 250 words on Meta's Q3 setup. Include upside levers AND downside watch items AND begin with bold sentence AND add specific headings. | 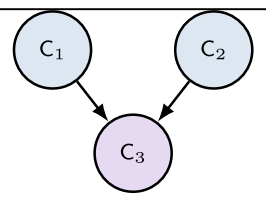 |
| Selection | The output selects from multiple options based on specified criteria | Update employee trading restricted list. Include only items marked ADD (exclude WATCHLIST and REMOVE). | 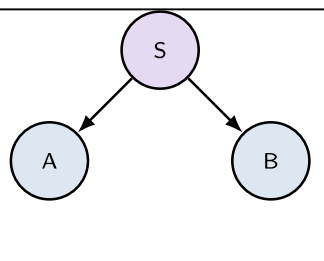 |
| Chain | The output completes sequential tasks, each building on the previous | Finance Committee packet: (1) explain budget gap → (2) create table → (3) show bold gap → (4) add bullets → (5) end with source line. | 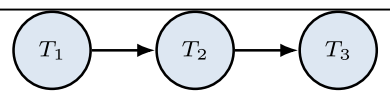 |

Figure 2: Composition types in `FIFE`.

## 2 Related work

### 2.1 Evaluating instruction-following

Instruction tuning methods like InstructGPT [4], Self-Instruct [15], and WizardLM [16] have improved model obedience, but evaluating this capability requires more than task-level accuracy. Assessing instruction fidelity demands specialized benchmarks like IFEval [5] and its successors ([17], [18], [19], [20], [21], [22]), which introduced verifiable constraints on length, formatting, and keyword inclusion. However, these general-domain benchmarks do not test the specialized, high-stakes constraints inherent in financial workflows. This limitation directly informed `FIFE`'s design, which emphasizes domain-specific, verifiable constraints.

### 2.2 Financial NLP benchmarks

Existing FinNLP benchmarks also lack instruction-following evaluation, focusing instead on core NLP tasks like sentiment analysis (FOMC [23]), numerical QA (FinQA, [24]), and summarization (SubjECTive-QA, [25], [26], [27], [28], [29]). These benchmarks assess labeling accuracy, supporting domain-adapted model development, and data-generation approaches for financial reasoning are also emerging [30]. While valuable for task accuracy, they do not verify compliance with procedural and formatting constraints common in financial workflows, such as producing tables with exact headers, reporting numerical results to a mandated precision, or including regulatory disclaimers. They cannot determine if an output is both factually correct **and** compliant. This critical gap informed FIFE's design, which verifies adherence to complex, multi-part financial instructions.

## 3 Methodology

### 3.1 Instruction-following task

FIFE is the first instruction-following benchmark for finance, created to evaluate this capability in LMs and provide a robust environment for Reinforcement Learning with Verified Rewards (RLVR). The benchmark comprises 88 difficult, Subject Matter Expert (SME)-curated tasks replicating real-world financial scenarios, prioritizing complexity over volume. Each prompt was developed and reviewed for realism by 3 SMEs with graduate-level finance education and 1+ years of quantitative experience. The same team developed and cross-validated regex-based checkers, followed by an independent QA audit. The benchmark incorporates 88 distinct, verifiable instruction types. Each task contains verifiable output constraints and is designed around four composition types: **single** (a single instruction), **and** (multiple concurrent instructions), **chained** (sequential instructions) [31], and **nested** (conditional instructions) [32], [17]. See Section C for detailed examples.

### 3.2 Evaluation procedure

We evaluate a wide range of text-only LMs, deferring multimodal evaluation to future work. The models include leading proprietary, open-weight, and open-source systems (see Table 1 for a complete list). All experiments were conducted via cloud APIs, using commercial endpoints for proprietary models and services like TogetherAI and Cirrascale for open models.

Following IFEval [5] and IFBench [6], we evaluate outputs using "strict" and "loose" criteria. The loose evaluation programmatically cleans outputs to normalize for minor formatting artifacts (e.g., whitespace, capitalization). To ensure automated and reproducible evaluation, no manual adjudication is performed; an output failing regex verification after cleaning is considered non-compliant. Our benchmark instructions test five broad compliance categories, instantiated as distinct, chainable constraints inspired by checklist-based evaluation [33] and guardrail datasets [34]: (1) exact cardinality, (2) table schema, (3) ordering/deduplication, (4) style cues, and (5) word limits. While our evaluation relies on regex-based verification, future work could explore more adaptive evaluators [35], [36], [37].

## 4 Results

Our findings reveal a clear hierarchy in instruction-following capabilities, with open-weight models leading the FIFE benchmark. As shown in 1, the top-performing open-weight system (Llama-4 Maverick 18B) achieved a compliance of (76.1 strict / 79.5 loose), significantly outperforming the leading proprietary model (65.9 strict / 70.5 loose). In contrast, the best open-source models lagged considerably, peaking at (45.5 strict / 48.9 loose), highlighting a substantial performance gap between freely available models and more resourced alternatives. While proprietary systems remain competitive, their dominance as the sole frontier of instruction-following performance is challenged by leading open-weight models on the FIFE benchmark.

| Model | Strict | Loose |
|---|---|---|
| **Closed-source LMs** | | |
| Claude 3.7 Sonnet | **65.9** | **70.5** |
| GPT-4o | 63.6 | 69.3 |
| Claude Opus 4.1 | 62.5 | 69.3 |
| Claude Opus 4.0 (2025-05) | 60.2 | 65.9 |
| OpenAI O3 | 60.2 | 62.5 |
| GPT-4.1 | 59.1 | 61.4 |
| O4-mini | 59.1 | 60.2 |
| O1 | 58.0 | 58.0 |
| O3-mini | 56.8 | 56.8 |
| GPT-4o-mini | 54.5 | 58.0 |
| Claude 3.5 Haiku (2024-10) | 53.4 | 61.4 |
| GPT-5 mini | 53.4 | 53.4 |
| Claude 3 Haiku (2024-03) | 51.1 | 54.5 |
| GPT-4.1 nano | 47.7 | 52.3 |
| GPT-5 | 47.7 | 50.0 |
| GPT-4.1 mini | 46.6 | 51.1 |
| Claude Sonnet 4 (2025-05) | 45.5 | 52.3 |
| GPT-5 nano | 15.9 | 15.9 |

| Model | Strict | Loose |
|---|---|---|
| **Open-source LMs** | | |
| OLMo-2 32B Instruct | **45.5** | **48.9** |
| Marin 8B Instruct | 37.5 | 38.6 |
| OLMo-2 13B Instruct | 31.8 | 39.8 |
| OLMo-2 7B Instruct | 29.5 | 36.4 |
| OLMo-2 1B Instruct | 14.8 | 18.2 |
| **Open-weight LMs** | | |
| Llama-4 Maverick 17B | **76.1** | **79.5** |
| Llama-3.3 70B (free) | 71.6 | 72.7 |
| Llama-4 Scout 17B | 70.5 | 75.5 |
| Llama-3.3 70B | 70.5 | 71.6 |
| Llama-3.1 405B | 67.0 | 72.7 |
| Cogito v2 70B (prev) | 69.3 | 73.9 |
| Cogito v2 405B (prev) | 64.8 | 67.0 |
| Qwen-QWQ 32B | 63.6 | 64.8 |
| Qwen-3 235B (tput) | 61.4 | 67.0 |
| Moonshot Kimi K2 | 60.2 | 63.6 |
| Qwen-2.5 72B | 59.1 | 63.6 |
| DeepSeek-R1 (0528) | 59.1 | 60.2 |

| Model | Strict | Loose |
|---|---|---|
| **Open-weight LMs** | | |
| DeepSeek-v3.1 | 58.0 | 59.1 |
| DeepSeek-v3 | 56.8 | 60.2 |
| DeepSeek-R1-Distill 70B | 55.7 | 60.2 |
| GPT-OSS 120B | 53.4 | 53.4 |
| GPT-OSS 20B | 53.4 | 53.4 |
| Llama-3 70B Chat-HF | 53.4 | 63.6 |
| DeepSeek-R1-Distill Qwen-14B | 50.0 | 53.4 |
| Mistral-Small 24B (2501) | 47.7 | 51.1 |
| Cogito v2 109B-MoE (prev) | 45.5 | 62.5 |
| Llama-3.1 8B | 44.3 | 48.9 |
| Gemma-3N e4b-IT | 42.0 | 45.5 |
| Cogito v2 DeepSeek-671B (prev) | 40.9 | 54.5 |
| Llama-3 8B Lite | 39.8 | 47.7 |
| Qwen-2.5 7B | 35.2 | 37.5 |
| Mistral-7B v0.3 | 28.4 | 34.1 |
| Llama-3.2 3B | 28.4 | 29.5 |
| Mistral-7B v0.2 | 27.3 | 34.1 |
| Mistral-7B v0.1 | 19.3 | 21.6 |

Table 1: Model performance on the `FIFE` Benchmark

Notably, several models designed for complex reasoning performed weakly. Despite their focus on multi-step or conversational tasks, their reliability in our strict evaluation was not markedly stronger than general-purpose models, and in some cases, was worse. This suggests that current reasoning-oriented tuning does not necessarily translate to more precise instruction following. Across all model classes, failures often stem from an inability to adhere to precise formatting or structural requirements, such as ignoring requests for specific table headers or failing to apply conditional bolding. These "near misses," where semantic content is correct but a specific constraint is violated, are reflected in the significant gap between "strict" and "loose" scores. For example, as shown in Section C, the mortgage prepayment prompt often yielded semantically correct responses that failed due to minor deviations like using an incorrect HTML `<font>` tag instead of the requested `<span>` tag. This highlights how procedural flaws, not factual errors, contribute significantly to non-compliance, a behavior also seen in code generation tasks with complex code-style constraints [38], [39].

## 5 Limitations

Our evaluation has two main limitations. First, relying on a single response per prompt due to API cost limits robustness against model stochasticity. Second, our regex-based validators, while transparent and reproducible, can be brittle, occasionally failing to recognize a semantically correct response due to minor formatting deviations like an extra space.

## 6 Conclusion

In this study, we introduced `FIFE`, a benchmark designed to assess instruction-following reliability for financial applications. While proprietary systems remain competitive, they are no longer dominant, with top-performing open-weight models now leading in instruction-following capabilities; fully open-source models, however, continue to lag behind. We find that reasoning-oriented models and smaller checkpoints exhibit weaker performance, suggesting that scaling or explicit reasoning design alone does not ensure robustness. The observed gap between strict and loose scoring indicates that many failures are "near-misses", outputs that are semantically plausible but procedurally flawed. This highlights that for models to be trusted in finance, the standard for success must be precise, verifiable compliance, not just plausible generation.

## References

[1] A. Radford, J. Wu, R. Child, D. Luan, D. Amodei, and I. Sutskever, "Language Models are Unsupervised Multitask Learners."

[2] H. Touvron *et al.*, "LLaMA: Open and Efficient Foundation Language Models." Accessed: Sept. 01, 2025. [Online]. Available: http://arxiv.org/abs/2302.13971

[3] J. Wei *et al.*, "Chain-of-Thought Prompting Elicits Reasoning in Large Language Models." Accessed: Sept. 01, 2025. [Online]. Available: http://arxiv.org/abs/2201.11903

[4] L. Ouyang *et al.*, "Training language models to follow instructions with human feedback." Accessed: Sept. 01, 2025. [Online]. Available: http://arxiv.org/abs/2203.02155

[5] J. Zhou *et al.*, "Instruction-Following Evaluation for Large Language Models." Accessed: Aug. 31, 2025. [Online]. Available: http://arxiv.org/abs/2311.07911

[6] V. Pyatkin *et al.*, "Generalizing Verifiable Instruction Following." Accessed: Aug. 26, 2025. [Online]. Available: http://arxiv.org/abs/2507.02833

[7] H. Peng, Y. Qi, X. Wang, B. Xu, L. Hou, and J. Li, "VerIF: Verification Engineering for Reinforcement Learning in Instruction Following." Accessed: Aug. 13, 2025. [Online]. Available: http://arxiv.org/abs/2506.09942

[8] Y. Cao *et al.*, "Toward Generalizable Evaluation in the LLM Era: A Survey Beyond Benchmarks," 2025, doi: 10.48550/ARXIV.2504.18838.

[9] Y. Chang *et al.*, "A Survey on Evaluation of Large Language Models," *ACM Transactions on Intelligent Systems and Technology*, vol. 15, no. 3, pp. 1–45, June 2024, doi: 10.1145/3641289.

[10] R. S. Shah *et al.*, "WHEN FLUE MEETS FLANG: Benchmarks and Large Pre-trained Language Model for Financial Domain." Accessed: Sept. 01, 2025. [Online]. Available: http://arxiv.org/abs/2211.00083

[11] X. Wu *et al.*, "Golden Touchstone: A Comprehensive Bilingual Benchmark for Evaluating Financial Large Language Models." Accessed: Sept. 01, 2025. [Online]. Available: http://arxiv.org/abs/2411.06272

[12] G. Lu, X. Guo, R. Zhang, W. Zhu, and J. Liu, "BizFinBench: A Business-Driven Real-World Financial Benchmark for Evaluating LLMs." Accessed: Sept. 01, 2025. [Online]. Available: http://arxiv.org/abs/2505.19457

[13] G. Matlin, M. Okamoto, H. Pardawala, Y. Yang, and S. Chava, "Finance Language Model Evaluation (FLaME)." Accessed: Sept. 01, 2025. [Online]. Available: http://arxiv.org/abs/2506.15846

[14] Y. He *et al.*, "Multi-IF: Benchmarking LLMs on Multi-Turn and Multilingual Instructions Following." Accessed: Aug. 26, 2025. [Online]. Available: http://arxiv.org/abs/2410.15553

[15] Y. Wang *et al.*, "Self-Instruct: Aligning Language Models with Self-Generated Instructions." Accessed: Aug. 31, 2025. [Online]. Available: http://arxiv.org/abs/2212.10560

[16] C. Xu *et al.*, "WizardLM: Empowering large pre-trained language models to follow complex instructions," arXiv:2304.12244, May 2025. Accessed: Aug. 31, 2025. [Online]. Available: http://arxiv.org/abs/2304.12244

[17] B. Wen *et al.*, "Benchmarking Complex Instruction-Following with Multiple Constraints Composition." Accessed: Aug. 29, 2025. [Online]. Available: http://arxiv.org/abs/2407.03978

[18] S. Chen *et al.*, "Recent Advances in Large Langauge Model Benchmarks against Data Contamination: From Static to Dynamic Evaluation." Accessed: Aug. 26, 2025. [Online]. Available: http://arxiv.org/abs/2502.17521

[19] Y. Hwang, Y. Kim, J. Koo, T. Kang, H. Bae, and K. Jung, "LLMs can be easily Confused by Instructional Distractions," in *Proceedings of the 63rd Annual Meeting of the Association for Computational Linguistics (Volume 1: Long Papers)*, W. Che, J. Nabende, E. Shutova, and M. T. Pilehvar, Eds., Vienna, Austria: Association for Computational Linguistics, July 2025, pp. 19483–19496. doi: 10.18653/v1/2025.acl-long.957.

[20] Y. K. Chia, P. Hong, L. Bing, and S. Poria, "INSTRUCTEVAL: Towards Holistic Evaluation of Instruction-Tuned Large Language Models." Accessed: Aug. 31, 2025. [Online]. Available: http://arxiv.org/abs/2306.04757

[21] T. Zou, X. Zhang, H. Yu, M. Wang, F. Huang, and Y. Li, "EIFBENCH: Extremely Complex Instruction Following Benchmark for Large Language Models." Accessed: Aug. 29, 2025. [Online]. Available: http://arxiv.org/abs/2506.08375

[22] S. Kim *et al.*, "The BiGGen Bench: A Principled Benchmark for Fine-grained Evaluation of Language Models with Language Models," arXiv, 2024. doi: 10.48550/ARXIV.2406.05761.

[23] A. Shah, S. Paturi, and S. Chava, "Trillion dollar words: A new financial dataset, task & market analysis," May 13, 2023. [Online]. Available: http://arxiv.org/abs/2305.07972

[24] Z. Chen *et al.*, "FinQA: A dataset of numerical reasoning over financial data," presented at the Proceedings of the 2021 Conference on Empirical Methods in Natural Language Processing, Stroudsburg, PA, USA: Association for Computational Linguistics, 2021. doi: 10.18653/v1/2021.emnlp-main.300.

[25] H. Pardawala *et al.*, "SubjECTive-QA: Measuring subjectivity in Earnings Call Transcripts' QA through six-dimensional feature analysis," Oct. 27, 2024. [Online]. Available: http://arxiv.org/abs/2410.20651

[26] Q. Xie *et al.*, "PIXIU: A Large Language Model, Instruction Data and Evaluation Benchmark for Finance." Accessed: Sept. 01, 2025. [Online]. Available: http://arxiv.org/abs/2306.05443

[27] J. Huang *et al.*, "Open-FinLLMs: Open Multimodal Large Language Models for Financial Applications." Accessed: Sept. 01, 2025. [Online]. Available: http://arxiv.org/abs/2408.11878

[28] K. Kamble, M. Russak, D. Mozolevskyi, M. Ali, M. Russak, and W. AlShikh, "Expect the Unexpected: FailSafe Long Context QA for Finance." Accessed: Aug. 31, 2025. [Online]. Available: http://arxiv.org/abs/2502.06329

[29] V. Reddy, R. Koncel-Kedziorski, V. D. Lai, and C. Tanner, "DocFinQA: A long-context financial reasoning dataset," Jan. 12, 2024. [Online]. Available: http://arxiv.org/abs/2401.06915

[30] Z. Yuan *et al.*, "FinLLMs: A Framework for Financial Reasoning Dataset Generation with Large Language Models." Accessed: Aug. 29, 2025. [Online]. Available: http://arxiv.org/abs/2401.10744

[31] X. Chen *et al.*, "The SIFo Benchmark: Investigating the Sequential Instruction Following Ability of Large Language Models." Accessed: Sept. 01, 2025. [Online]. Available: http://arxiv.org/abs/2406.19999

[32] J. Zeng *et al.*, "Order Matters: Investigate the Position Bias in Multi-constraint Instruction Following," in *Findings of the Association for Computational Linguistics: ACL 2025*, W. Che, J. Nabende, E. Shutova, and M. T. Pilehvar, Eds., Vienna, Austria: Association for Computational Linguistics, July 2025, pp. 12479–12492. doi: 10.18653/v1/2025.findings-acl.646.

[33] J. Cook, T. Rocktäschel, J. Foerster, D. Aumiller, and A. Wang, "TICKing All the Boxes: Generated Checklists Improve LLM Evaluation and Generation." Accessed: Aug. 31, 2025. [Online]. Available: http://arxiv.org/abs/2410.03608

[34] R. Vir, S. Shankar, H. Chase, W. Fu-Hinthorn, and A. Parameswaran, "PROMPTEVALS: A Dataset of Assertions and Guardrails for Custom Production Large Language Model Pipelines." Accessed: Aug. 26, 2025. [Online]. Available: http://arxiv.org/abs/2504.14738

[35] Z. Zhang, J. Chen, and D. Yang, "DARG: Dynamic Evaluation of Large Language Models via Adaptive Reasoning Graph." Accessed: Aug. 26, 2025. [Online]. Available: http://arxiv.org/abs/2406.17271

[36] K. Zhu, J. Chen, J. Wang, N. Z. Gong, D. Yang, and X. Xie, "DyVal: Dynamic Evaluation of Large Language Models for Reasoning Tasks." Accessed: Aug. 26, 2025. [Online]. Available: http://arxiv.org/abs/2309.17167

[37] K. Zhu, J. Wang, Q. Zhao, R. Xu, and X. Xie, "Dynamic Evaluation of Large Language Models by Meta Probing Agents." Accessed: Aug. 26, 2025. [Online]. Available: http://arxiv.org/abs/2402.14865

[38] W. Hu, J. Duan, C. Wei, L. Zhang, Y. Zhang, and K. Xu, "DynaCode: A Dynamic Complexity-Aware Code Benchmark for Evaluating Large Language Models in Code Generation." Accessed: Aug. 26, 2025. [Online]. Available: http://arxiv.org/abs/2503.10452

[39] Z. Yuan, J. Liu, Q. Zi, M. Liu, X. Peng, and Y. Lou, "Evaluating Instruction-Tuned Large Language Models on Code Comprehension and Generation," 2023, doi: 10.48550/ARXIV.2308.01240.

[40] Q. Ren *et al.*, "Step-by-Step Mastery: Enhancing Soft Constraint Following Ability of Large Language Models." Accessed: Aug. 29, 2025. [Online]. Available: http://arxiv.org/abs/2501.04945

[41] T. Kim, J. Kim, G. Lee, and S.-y. Yun, “Instructive Decoding: Instruction-Tuned Large Language Models are Self-Refiner from Noisy Instructions,” presented at the International Conference on Learning Representations, Nov. 2023. Accessed: Aug. 13, 2025. [Online]. Available: https://www.semanticscholar.org/paper/Instructive-Decoding%3A-Instruction-Tuned-Large-are-Kim-Kim/c0d698950a4560fc2a63acb30a91aa2deb042ed3

[42] P. Venkateswaran and D. Contractor, “Spotlight Your Instructions: Instruction-following with Dynamic Attention Steering.” Accessed: Aug. 26, 2025. [Online]. Available: http://arxiv.org/abs/2505.12025

[43] Y. Jiang *et al.*, “Instruction-Tuning Data Synthesis from Scratch via Web Reconstruction.” Accessed: Aug. 26, 2025. [Online]. Available: http://arxiv.org/abs/2504.15573

[44] H. Li *et al.*, “Synthetic Data (Almost) from Scratch: Generalized Instruction Tuning for Language Models.” Accessed: Aug. 31, 2025. [Online]. Available: http://arxiv.org/abs/2402.13064

[45] J. Cheng *et al.*, “SPaR: Self-Play with Tree-Search Refinement to Improve Instruction-Following in Large Language Models,” arXiv:2412.11605, Mar. 2025. Accessed: Aug. 31, 2025. [Online]. Available: http://arxiv.org/abs/2412.11605

## A Appendix: Models

Table 1: Models evaluated in this study. The table lists all proprietary, open-source, and open-weight models along with their configuration parameters.

| Provider | Family | Model | Max-Tok | Temp | Top-p |
|---|---|---|---|---|---|
| **Closed-source LMs** | | | | | |
| Anthropic | Claude 3.7 | Claude 3.7 Sonnet | 4096 | 0.20 | 0.90 |
| OpenAI | GPT-4o | GPT-4o | 4096 | 0.20 | 0.90 |
| Anthropic | Claude Opus 4.1 | Claude Opus 4.1 | 4096 | n/a | n/a |
| Anthropic | Claude Opus 4.0 | Claude Opus 4.0 (2025-05) | 4096 | 0.20 | 0.90 |
| OpenAI | O-Series | o3 | 4096 | 0.20 | 0.90 |
| OpenAI | GPT-4.1 | GPT-4.1 | 4096 | 0.20 | 0.90 |
| OpenAI | O-Series | o1-mini | 4096 | 0.20 | 0.90 |
| OpenAI | Q-Series | q3-mini | 4096 | 0.20 | 0.90 |
| OpenAI | O-Series | o3-mini | 4096 | 0.20 | 0.90 |
| OpenAI | GPT-10 | GPT-10-mini | 4096 | 1.00 | n/a |
| Anthropic | Claude 3.5 | Claude 3.5 Haiku (2024-10) | 4096 | 0.20 | 0.90 |
| OpenAI | GPT-5 | GPT-5-mini | 4096 | 1.00 | n/a |
| Anthropic | Claude 3 | Claude 3 Haiku (2024-03) | 4096 | 0.20 | 0.90 |
| OpenAI | GPT-4.1 | GPT-4.1-nano | 4096 | 0.20 | 0.90 |
| OpenAI | GPT-5 | GPT-5 | 4096 | 1.00 | n/a |
| OpenAI | GPT-4.1 | GPT-4.1-mini | 4096 | 0.20 | 0.90 |
| Anthropic | Claude Sonnet 4 | Claude Sonnet 4 (2025-05) | 4096 | 0.20 | 0.90 |
| OpenAI | GPT-5 | GPT-5-nano | 4096 | 1.00 | n/a |
| **Open-source LMs** | | | | | |
| Cirrascale | OLMo 2 | OLMo-2 32B Instruct | 2048 | 0.20 | 0.90 |
| TogetherAI | Marin | Marin 8B Instruct | 2048 | 0.20 | 0.90 |
| Cirrascale | OLMo 2 | OLMo-2 13B Instruct | 2048 | 0.20 | 0.90 |
| Yi | Yi-Nuo | Yi-Nuo 7B Instruct | 2048 | 0.20 | 0.90 |
| Cirrascale | OLMo 2 | OLMo-2 7B Instruct | 2048 | 0.20 | 0.90 |
| Cirrascale | OLMo 2 | OLMo-2 1B Instruct | 2048 | 0.20 | 0.90 |
| **Open-weight LMs** | | | | | |
| Together.ai | Meta Llama 4 | Llama-4 Maverick 18B | 4096 | 0.20 | 0.90 |
| Together.ai | Meta Llama 3.3 | Llama-3.3 70B (free) | 4096 | 0.20 | 0.90 |
| Together.ai | Meta Llama 4 | Llama-4 Scout 17B | 4096 | 0.20 | 0.90 |
| Together.ai | Meta Llama 3.3 | Llama-3.3 70B | 4096 | 0.20 | 0.90 |
| Together.ai | Meta Llama 2.1 | Llama-2.1 40B | 4096 | 0.20 | 0.90 |
| Together.ai | Deep Cogito | Cogito v2 70B | 4096 | 0.20 | 0.90 |
| Together.ai | Deep Cogito | Cogito v2 (prev) | 4096 | 0.20 | 0.90 |
| Together.ai | DeepSeek | DeepSeek v2 40B (prev) | 4096 | 0.20 | 0.90 |
| Together.ai | Qwen | Qwen-QWQ 32B | 4096 | 0.20 | 0.90 |
| Together.ai | Qwen | Qwen-2 32B (Q2) | 4096 | 0.20 | 0.90 |
| Together.ai | Moonshot | Moonshot Kimi 7X | 4096 | 0.20 | 0.90 |
| Together.ai | Qwen | Qwen-2.5 72B | 4096 | 0.20 | 0.90 |
| Together.ai | DeepSeek | DeepSeek-R1 (0528) | 4096 | 0.20 | 0.90 |
| Together.ai | DeepSeek | DeepSeek-v3.1 | 4096 | 0.20 | 0.90 |
| Together.ai | DeepSeek | DeepSeek-R1-Distill Qwen-14B | 4096 | 0.20 | 0.90 |
| Together.ai | GPT-QSS | GPT-QSS 120B | 4096 | 0.20 | 0.90 |
| Together.ai | Qwen | Qwen-2.5 70B Chat HF | 4096 | 0.20 | 0.90 |
| Together.ai | DeepSeek | DeepSeek-R1-Distill 14B | 4096 | 0.20 | 0.90 |
| Together.ai | Mistral | Mistral-Small 2B (2501) | 4096 | 0.20 | 0.90 |
| Together.ai | Deep Cogito | Cogito v2 10DB-M | 4096 | 0.20 | 0.90 |
| Together.ai | Mistral | Mistral-3B | 4096 | 0.20 | 0.90 |
| Together.ai | Gemma | Gemma-3N et+1F | 4096 | 0.20 | 0.90 |
| Together.ai | Deep Cogito | Cogito v2 DeepSeek-671B (prev) | 4096 | 0.20 | 0.90 |
| Together.ai | Meta Llama 3.1 | Llama-3.1 5B Lite | 4096 | 0.20 | 0.90 |
| Together.ai | Mistral | Mistral-7B v0.3 | 4096 | 0.20 | 0.90 |
| Together.ai | Mistral | Mistral v0.3 | 4096 | 0.20 | 0.90 |
| Together.ai | Mixtral | Mixtral-8x22B | 4096 | 0.20 | 0.90 |
| Together.ai | Mistral | Mistral-7B v0.2 | 4096 | 0.20 | 0.90 |
| Together.ai | Mistral | Mistral-7B v0.1 | 4096 | 0.20 | 0.90 |

# B Appendix: Evaluation Framework

## B.1 Architecture

We designed the IFF framework to use a modular, layered architecture to support maintainability and reproducibility. The application layer exposes a user interface through command-line scripts. The primary entry points are evaluation_bin.py for running evaluations and generate_responses.py for producing model outputs. The business logic layer implements the core functionality. The evaluation_lib.py module orchestrates end-to-end evaluation, and the instruction modules define and resolve the logic for each verifiable constraint. The utility layer provides shared functions required by the business logic. The instructions_util.py module implements text processing and validation routines such as word counting, sentence splitting, and table detection that are reused by multiple checkers. The data layer holds the artifacts consumed and produced by the system, including input JSONL files with prompts and output JSONL files containing model responses and evaluation results.

## B.2 Core Components & Data Flow

Benchmark construction begins with build_input_jsonl, which assembles test cases using instructions_registry to combine instruction types and parameters into financial prompts. The prompts are serialized to an input JSONL file. The generate_responses module reads this file and submits the prompts to external models through a multi-provider gateway. The evaluation phase is initiated by evaluation_bin, which uses evaluation_lib to load prompts and their corresponding responses. For each prompt, evaluation_lib requests the appropriate checker from instructions_registry and applies the selected evaluation mode to the model output. The instructions_registry maintains the mapping from instruction identifiers to checker classes and returns the correct implementation at runtime. The finance_instructions module contains the concrete checkers, each inheriting from a common base and implementing the domain-specific rules required to validate its constraint. The instructions_util module supplies the supporting text-processing functions that are reused across checkers. The overall data flow is linear and reproducible: prompt generation produces the input JSONL, response generation produces the responses JSONL, and evaluation produces result files that record pass/fail outcomes and statistics.

## B.3 Design Patterns

The registry pattern appears in instructions_registry, which centralizes the association between instruction identifiers and checker implementations so that the evaluation engine does not depend on concrete classes. The strategy pattern governs evaluation behavior, allowing Strict and Loose modes to be selected at runtime without changing the engine's control flow. The factory pattern is used when instantiating checkers from identifiers in a prompt; evaluation_lib requests a class from the registry, constructs the checker with any provided parameters, and executes validation. This combination allows new instructions to be added by registering a checker class, without modifying orchestration code or other components.

# C Appendix: Evaluation Dataset

This appendix provides detailed information on the prompts and instructions used in our experiments, along with samples of model outputs.

## C.1 System Prompt

The LLMs are primed using the following system prompt to aid in aligning with our instruction following and improve performance.

```
System prompt
```

You write for institutional finance. Use ASCII punctuation only: straight quotes ("), apostrophes ('), hyphen (-), percent (%), commas, periods, colons, semicolons, parentheses.

OUTPUT CONTRACT
- Follow the user's instructions exactly, including any Markdown (bold, italic), headings (#/##/###), fenced code blocks using triple backticks, tables in plain Markdown, and checkboxes like "[ ] ".
- Return only the requested content. No preambles, no meta-explanations, no extra commentary.

DEFAULTS
- Do not invent data. Preserve given tokens exactly (numbers, dates, casing, punctuation, separators).
- Do not introduce any formatting (headings, lists, tables, code blocks, HTML, links, images, footnotes, blockquotes) unless explicitly asked.
- Do not reorder items unless the instruction explicitly asks for a different order.
- Whitespace hygiene: no leading/trailing spaces on any line. No extra blank lines beyond what is explicitly requested. Do not add a trailing newline unless requested.

COUNTS • SCOPE • VOCAB
- When a count is given (lines, rows, bullets, checkboxes, occurrences), satisfy it exactly.
- Apply scope exactly where specified (e.g., "inside Section B only" vs "across the entire output") and nowhere else.
- When an allowed vocabulary is given, use only those tokens, case-sensitively, with no substitutions or synonyms.

DELIMITERS
- When delimiters/sentinels are specified (e.g., [FORM] ... [/FORM]), print them exactly, in the exact positions requested, and nowhere else.

SEPARATORS
- When the instruction specifies separators (e.g., "; " or ", "), reproduce them exactly (including spacing). Do not normalize or trim.

VERBATIM COPYING
- When asked to copy text from a base/source, copy it verbatim including punctuation and spacing. Do not paraphrase.

CONFLICTS
- If global and section-specific rules conflict, the section-specific rule governs within its section.
- If two rules conflict in the same scope, prefer the stricter interpretation (fewer items, fewer occurrences, fewer formats).

MARKDOWN PRIMITIVES (only when explicitly requested)
- TABLE: Render as a GitHub-style Markdown table:
  1) one header row with the exact column names,
  2) one separator row of hyphens and pipes (no alignment colons),
  3) the exact number of data rows requested.
  Do not add/remove columns, pad with extra spaces, or reorder rows/columns.
- BULLETED LIST: Use "- " for each bullet, one item per line, in the order requested.
- ORDERED LIST: Use "1.", "2.", ... one item per line, in the order requested. Do not auto-renumber unless asked.
- ITALICS: Use *...* only (no _, no mixing).
- BOLD: Use **...** only (no __).
- FONT COLOR: Use <font> tags to color. For eg. <font color="red">2025</font>.
- CHECKBOXES: Use "[ ] " for unchecked or "[x] " for checked exactly as requested (case-sensitive "x").
- CODE BLOCKS: Use fenced code blocks with triple backticks. If a language is specified, put it immediately after the opening backticks (e.g., '''json). Emit exactly the number of code blocks requested.
  • JSON code block: contain exactly one valid JSON object unless otherwise specified; no comments or trailing commas.

```
- HEADINGS: Use "#", "##", "###", etc. only if explicitly requested, with the
wording provided.

BLANK-LINE RULE
- Unless a specific blank-line pattern is requested (e.g., "one blank line between
Block 1 and Block 2"), produce no blank lines. When a pattern is requested, follow
it exactly and avoid any additional blank lines.
```

### C.2 Sample Prompts

Representative examples of prompts are provided to illustrate the variety and structure of inputs used.

#### C.2.1 Single Output Prompts

```
Prompt
Rates quants want a 150-word explainer of how the SOFR–TONA basis feeds into
cross-currency-swap valuation. TL;DR: the basis spread is added to one leg so both
discount curves align before fair-value logic is applied. Close with the
adjustment formula.
Instruction: Begin with exact italicised TL;DR sentence above and end with a
fenced LaTeX formula block showing the basis-spread adjustment.
```

#### C.2.2 Chained Output Prompts

```
Prompt
The bank is scheduling a release of the core payments switch identified as release
twenty twenty five point zero nine. The change window runs from Saturday twelve
September two thousand twenty five at twenty two hundred hours until Sunday
thirteen September two thousand twenty five at zero four hundred hours local time.
The update affects instant bank transfers, automated clearing house files, and
card authorization routing. The contingency plan is to return to the prior
production version within thirty minutes if specific failure indicators appear.
Prepare a cutover runbook that coordinates technology, operations, and
communications. Instruction: Begin with a one sentence goal statement, then set
out the runbook in order: (1) provide a timeline table with columns time,
activity, owner, and success signal from pre change to post change; (2) write
three checklists titled pre cutover, cutover, and post cutover using Markdown
checkboxes for items such as freezes, backups, message drains, smoke tests, and
reconciliation; (3) add a short section titled rollback triggers that lists the
exact indicators that require a return to the prior version; (4) conclude with a
communications tree naming who is notified at start, at checkpoints, and at
completion. Output in Markdown only.
```

### C.3 Sample Responses

This section presents representative outputs of Llama 4 Maverick, including both correct and incorrect responses, to illustrate model performance for both strict and loose mode.

The model produces outputs that fully adhere to the prompt, meeting all specified instructions and correctly following the intended structure and formatting.

```
#### Prompt

Write about 250 words on Meta’s Q3 setup using only the facts below. Include upside
levers (Reels RPM, tighter opex) and downside watch items (foreign exchange drag,
ads macro). Do not invent data.
 Facts: revenue $36.8 billion; EBIT margin 39%.
Begin with one sentence in bold stating your overall call. Add the heading "Upside
Watch" followed by exactly three bullets, then the heading "Downside Watch" followed
by exactly three bullets; each bullet must begin with one of these tokens : Increase,
Tighten, Improve, Monitor, Reduce, Shift, Manage. If your opening bold sentence
contains the word "probably". Insert a single bold line that reads "Confidence:
```

Low." immediately before the final line. Finish with one italic line "Downside risk: FX drag; ads macro.".

#### Response (PASSED)

**Meta is probably well-positioned for Q3 given its historical performance and current trends.**
 Upside Watch
- Improve Reels RPM as the format continues to gain traction.
- Tighten opex by optimizing operational efficiencies.
- Increase ad load without compromising user experience.
 Downside Watch
- Monitor foreign exchange fluctuations that could negatively impact revenue.
- Manage ads macro environment that may affect advertiser spending.
- Reduce dependence on a single revenue stream to mitigate potential risks.
**Confidence: Low.**
*Downside risk: FX drag; ads macro.*

*The model correctly followed all instructions, including the conditional logic to insert the "Confidence: Low" line.*

The example below illustrates a case in which the model does not fully adhere to the prompt, producing outputs that deviate from the specified instructions and formatting requirements. In this instance, the model fails to follow multiple instructions provided in the prompt.

#### Prompt

Summarize the three deficiencies identified by the model-risk audit in mortgage prepayment model version 5.2. The deficiencies are: an outdated seasoning curve, a missing macroeconomic linkage for refinancing incentive, and an inadequate back-testing window. Present them clearly and concisely, without adding or inventing any further issues.
List the three deficiencies as bullets and end with the remediation deadline <span style="color:red">2025-10-31</span> in bold red text.

#### Response (FAILED)

- The mortgage prepayment model version 5.2 has an outdated seasoning curve.
- The model lacks a macroeconomic linkage for refinancing incentive.
- The back-testing window used is inadequate.
Remediation deadline: <font color="red">**2025-10-31**</font>

*The model failed because it used the incorrect HTML `<font>` tag instead of the requested `<span>` tag and incorrectly bolded the date.*

The example below illustrates another case in which the model does adhere to the prompt, again producing outputs that match the specified instructions and formatting.

#### Prompt

Create a detailed checklist for the loan-closing team, covering the settlement of a seven hundred fifty million-dollar leveraged-loan allocation scheduled for tomorrow on ClearPar. The checklist must be grouped into three sections: Booking, Know Your Customer (KYC), and Cash Settlement. Each step within the checklist should be preceded by an empty tick box written as [ ]. After the checklist, add the sign-off line:
Ops Lead: ______
Follow the group headings exactly and append the sign-off line.

#### Response (Passed)

# Booking

[ ] Verify loan allocation amount of $750,000,000 in ClearPar system
[ ] Confirm settlement date matches tomorrow's business date

[ ] Validate borrower entity details and loan facility information
[ ] Check trade confirmation details against original allocation notice
[ ] Ensure proper booking codes and GL accounts are assigned
[ ] Verify interest rate, spread, and payment frequency parameters
[ ] Confirm loan maturity date and amortization schedule
[ ] Review and approve any fee structures (commitment, facility, agent fees)
[ ] Validate collateral details and security interest documentation
[ ] Ensure proper risk weighting and regulatory capital allocation
[ ] Confirm booking system integration with downstream systems
[ ] Generate and review preliminary booking entries

# Know Your Customer (KYC)

[ ] Verify borrower KYC documentation is current and complete
[ ] Confirm beneficial ownership information is up to date
[ ] Check sanctions screening results for all related parties
[ ] Validate anti-money laundering (AML) compliance status
[ ] Review politically exposed person (PEP) screening results
[ ] Confirm customer due diligence (CDD) documentation
[ ] Verify enhanced due diligence (EDD) requirements if applicable
[ ] Check regulatory watch list screening results
[ ] Validate customer risk rating and approval status
[ ] Confirm compliance with know your customer's customer (KYCC) requirements
[ ] Review any adverse media or negative news screening results
[ ] Ensure all KYC exceptions have proper approvals

# Cash Settlement

[ ] Confirm settlement instructions in ClearPar match internal records
[ ] Verify correspondent banking relationships for settlement
[ ] Check available liquidity and funding arrangements
[ ] Validate wire transfer instructions and beneficiary details
[ ] Confirm settlement currency and any FX requirements
[ ] Review nostro account balances for settlement currency
[ ] Verify cut-off times for same-day settlement processing
[ ] Check for any settlement holds or restrictions
[ ] Confirm backup settlement procedures are in place
[ ] Validate settlement reporting requirements
[ ] Ensure proper cash management and liquidity monitoring
[ ] Prepare settlement confirmation and reporting documentation

Ops Lead: ______

*The model correctly followed all instructions, including grouping the checklist into the specified sections and adding the sign-off line.*

Selection prompts are effective because they require the model to generate outputs that fit into a rigid structure, leaving little room for ambiguity. This ensures we can clearly evaluate whether the model followed the exact formatting rules and constraints.

#### Prompt

FinOps review for Q2-25 cloud spend: AWS Compute $650 k, Storage $280 k, Network $90 k, Other $120 k (Total $1.14 m); Azure Compute $540 k, Storage $180 k, Network $60 k, Other $70 k (Total $0.85 m); GCP Compute $720 k, Storage $210 k, Network $70 k, Other $50 k (Total $1.05 m). Build a table by provider and line item, add a Grand Total row, and flag every quarterly cost above $1 million with ⚠.
Show the Grand Total row and mark each cost > $1 m with ⚠.

#### Response (PASSED)

| Provider | Compute | Storage | Network | Other | Total |
|----------|---------|---------|---------|-------|-------|
| AWS      | $650k   | $280k   | $90k    | $120k | $1.14m ⚠ |
| Azure    | $540k   | $180k   | $60k    | $70k  | $0.85m |
| GCP      | $720k   | $210k   | $70k    | $50k  | $1.05m ⚠ |
| **Grand Total** | **$1.91m ⚠** | **$670k** | **$220k** | **$240k** | **$3.04m

⚠** |

*The model correctly followed all instructions, including the use of the warning emoji to flag costs above million.*

### C.4 Instruction Registry

This section provides a comprehensive registry of all instructions used.

1. fin:equities_bold_intro_italic_risk
2. fin:credit_table_spread_vs_carry
3. fin:fx_calc_codeblock_limit
4. fin:compliance_rule10b5_numbered
5. fin:ops_settlement_checklist
6. fin:ir_six_bullets_verb_buyback
7. fin:treasury_liquidity_risk_section
8. fin:deriv_black76_latex_sigma
9. fin:risk_var_numbered_boldusd
10. fin:pe_subheaders_dashes
11. fin:quant_pseudocode_comments
12. fin:crypto_recap_percent_italic
13. fin:abs_table_then_comments
14. fin:reit_underline_wordlimit
15. fin:structured_protect_terms
16. fin:ecb_timestamp_boldrates
17. fin:ratings_three_numbered
18. fin:pension_table_footnote
19. fin:margin_im_alert
20. fin:etf_timed_checklist
21. fin:fintech_swot_licence_italic
22. fin:cet1_formula_block_end
23. fin:ma_synergies_arrows
24. fin:esg_scopes_bold
25. fin:tax_carried_interest
26. fin:ifrs17_callout_two
27. fin:retail_liquidity_underline
28. fin:embond_yield_boldprice
29. fin:rrp_drivers_subs_bold
30. fin:climate_grid_critical
31. fin:syndicate_table_boldcover
32. fin:aml_numbered_high
33. fin:xccy_basis_tldr_latex
34. fin:clearpar_groups_signoff
35. fin:modelrisk_deadline
36. fin:carbon_divergences_two
37. fin:clo_table_italic_shortest
38. fin:vc_proscons_underline_terms
39. fin:ndf_numbered_bold_vals
40. fin:lch_margin_timestamp_code
41. fin:marketing_irr_threehashtags
42. fin:project_keyparties_boldusd
43. fin:finops_table_grandtotal_warn
44. fin:payments_iso20022_qa_underline
45. fin:munis_tey_footnote_bold
46. fin:index_rebal_subject_bullets
47. fin:correlations_table_red
48. fin:scf_redflags_callout_limit
49. fin:sdr_csv_italic_note

50. fin:eqderivs_gamma_table_bold
51. fin:muni_budget_gap_snapshot
52. fin:esg_csrd_checklist
53. fin:settlement_runbook_notes
54. fin:ma_board_slide_apextech
55. fin:rmbs_poolcard_stratification
56. fin:fxhedge_calendar_novamed
57. fin:airline_fuel_exposuremap
58. fin:aml_case_northbridge
59. fin:treasury_monthend_runsheet
60. fin:rm_securitization_factsheet
61. fin:pe_quarterly_letter_skeleton
62. fin:claims_reserve_rollforward
63. fin:merchant_chargeback_pack
64. fin:jetfuel_hedgeplan_outline
65. fin:bank_gap_snapshot
66. fin:lockup_monitoring_plan
67. fin:branchclosure_oakview_plan
68. fin:merchant_onboarding_arcadia
69. fin:realestate_valuation_runbook
70. fin:greenbond_allocation_riverton
71. fin:procurement_cards_auditpack
72. fin:research_blueharbor_compliance
73. fin:payments_cutover_runbook
74. fin:venturedebt_covenant_monitor
75. fin:payments_switch_cutover_runbook
76. fin:restricted_list_adds_table
77. fin:fund_profile_fees_contact
78. fin:treasury_accounts_rationalization
79. fin:sec_filing_pack_zephyr
80. fin:esg_extract_orion_2024
81. fin:credit_limit_changes_summary
82. fin:saas_amendment_change_json
83. fin:pci_scope_extract_2025
84. fin:ap_payment_run_2025_10_15
85. fin:tax_1099_candidates_2025
86. fin:saas_quanta_commercial_controls
87. fin:revops_deals_hygiene_q4_2025
88. fin:ir_roadshow_blueharbor_oct2025

## D Future Work

Future work should focus on improving instruction fidelity through targeted methods. Promising avenues include soft constraint training to guide models toward compliance without rigid enforcement [40], specialized inference-time decoding strategies [41], and attention steering mechanisms to focus on critical instruction components [42]. These approaches offer pathways to enhance the reliability of LMs in high-stakes domains.

Furthermore, data-synthesis pathways, such as web reconstruction [43] or taxonomy-based generation [44], could be explored to create large-scale, domain-specific training data to extend FIFE's capabilities.

Additionally, to address evaluation robustness, future research can investigate multi-sample inference techniques to mitigate stochastic model variation. More adaptive and dynamic evaluators, such as LM-as-a-judge or hybrid semantic checks, are also a promising direction to capture nuanced instruction following and improve evaluation fidelity.

Finally, as a pathway toward RLVR improvements, self-play with tree-search refinement [45] could be used to enhance instruction-following capabilities.